\documentclass[letterpaper, 10pt, conference]{style/ieeeconf}    

\makeatletter
\let\NAT@parse\undefined
\makeatother

\usepackage{dblfloatfix}

\usepackage[numbers,sectionbib,sort&compress]{natbib}
\usepackage{bm}
\usepackage{gensymb}
\usepackage{graphicx}
\usepackage{amsmath}
\usepackage{amssymb}
\usepackage{algorithm}
\usepackage[noend]{algpseudocode}
\usepackage{subcaption}
\usepackage{amsfonts}
\usepackage{siunitx}
\usepackage{booktabs}
\usepackage{makecell}
\usepackage{multirow}
\usepackage{upgreek}
\usepackage[font=small]{caption}
\usepackage[export]{adjustbox}
\usepackage{tikz}
\usepackage{sidecap} \sidecaptionvpos{figure}{c}
\usepackage{hyperref}
\hypersetup{colorlinks,breaklinks,
linkcolor=[rgb]{0.5,0.,0.},
citecolor=[rgb]{0.000,0.427,0.173},
urlcolor=[rgb]{0.031,0.318,0.612}}

\usepackage[nameinlink]{cleveref}
\usepackage{color}
\definecolor{CommentPink}{rgb}{1,0.2,0.5}
\definecolor{CommentBlue}{rgb}{0,0,1}
\definecolor{CommentGreen}{rgb}{0,1,0}

\newcommand\edit[1]{{\color{black}#1}}

\usepackage[printonlyused,withpage,nolist,nohyperlinks]{acronym}
\begin{acronym}

\acro{2D}{two-dimensional}

\acro{3D}{three-dimensional}

\acro{AHRS}{attitude and heading reference system}

\acro{AUV}{autonomous underwater vehicle}

\acro{CPP}{Chinese Postman Problem}

\acro{DoF}{degree of freedom}
\acrodefplural{DoF}[DoFs]{degrees of freedom}

\acro{DVL}{Doppler velocity log}

\acro{FSM}{finite state machine}

\acro{IMU}{inertial measurement unit}

\acro{LBL}{Long Baseline}

\acro{MCM}{mine countermeasures}

\acro{MDP}{Markov decision process}
\acrodefplural{MDP}[MDPs]{Markov decision processes}

\acro{POMDP}{Partially Observable Markov Decision Process}
\acrodefplural{POMDP}[POMDPs]{Partially Observable Markov Decision Processes}

\acro{PRM}{Probabilistic Roadmap}
\acrodefplural{PRM}[PRM]{Probabilistic Roadmaps}

\acro{ROI}{region of interest}
\acrodefplural{ROI}[ROIs]{regions of interest}

\acro{ROS}{Robot Operating System}

\acro{ROV}{remotely operated vehicle}

\acro{RRT}{Rapidly-exploring Random Tree}
\acrodefplural{RRT}[RRTs]{Rapidly-exploring Random Trees}

\acro{SLAM}{Simultaneous Localisation and Mapping}

\acro{SSE}{sum of squared errors}

\acro{STOMP}{Stochastic Trajectory Optimization for Motion Planning}

\acro{TRN}{Terrain-Relative Navigation}

\acro{UAV}{unmanned aerial vehicle}

\acro{USBL}{Ultra-Short Baseline}

\acro{IPP}{informative path planning}

\acro{FoV}{field of view}
\acrodefplural{FoV}[FoVs]{fields of view}

\acro{CDF}{cumulative distribution function}

\acro{ML}{maximum likelihood}

\acro{RMSE}{Root Mean Squared Error}
\acro{MLL}{Mean Log Loss}

\acro{GP}{Gaussian Process}
\acrodefplural{GP}[GPs]{Gaussian Processes}

\acro{KF}{Kalman Filter}

\acro{IP}{Interior Point}
\acro{BO}{Bayesian Optimization}

\acro{SE}{squared exponential}

\acro{UI}{uncertain input}

\acro{MCL}{Monte Carlo Localisation}
\acro{AMCL}{Adaptive Monte Carlo Localisation}

\acro{SSIM}{Structural Similarity Index}

\acro{MAE}{Mean Absolute Error}
\acro{RMSE}{Root Mean Squared Error}
\acro{AUSE}{Area Under the Sparsification Error curve}

\end{acronym}

\IEEEoverridecommandlockouts                           
\title{\LARGE \bf
Volumetric Occupancy Mapping With Probabilistic \\ Depth Completion for Robotic Navigation}
\author{Marija Popovi\'{c}$^{*,1,2}$, Florian Thomas$^{*,1}$, Sotiris Papatheodorou$^{1}$, Nils Funk$^{1}$, \\ Teresa Vidal-Calleja$^{3}$, Stefan Leutenegger$^{1,4}$
\thanks{This research is supported by the ESPRC ORCA Robotics Hub (EP/R026173/1), EPSRC grant Aerial ABM (EP/N018494/1), Imperial College London (including President's Scholarship), and SLAMcore Ltd.
It is partially funded by the Deutsche Forschungsgemeinschaft (DFG, German Research Foundation) under Germany’s Excellence Strategy - EXC 2070 – 390732324.
*Equal contribution.
$^{1}$Smart Robotics Lab, Department of Computing, Imperial College London.
$^{2}$Cluster of Excellence PhenoRob, Institute of Geodesy and Geoinformation, University of Bonn.
$^{3}$Centre for Autonomous Systems, Faculty of Engineering and IT, University of Technology Sydney.
$^{4}$Smart Robotics Lab, Department of Informatics, Technical University of Munich.
\texttt{mpopovic@uni-bonn.de}.}%
}

\begin{document}

\maketitle
\thispagestyle{empty}
\pagestyle{empty}

\begin{abstract}
In robotic applications, a key requirement for safe and efficient motion planning is the ability to map obstacle-free space in unknown, cluttered 3D environments. However, commodity-grade RGB-D cameras commonly used for sensing fail to register valid depth values on shiny, glossy, bright, or distant surfaces, leading to missing data in the map. To address this issue, we propose a framework leveraging probabilistic depth completion as an additional input for spatial mapping. We introduce a deep learning architecture providing uncertainty estimates for the depth completion of RGB-D images. Our pipeline exploits the inferred missing depth values and depth uncertainty to complement raw depth images and improve the speed and quality of free space mapping. Evaluations on synthetic data show that our approach maps significantly more correct free space with relatively low error when compared against using raw data alone in different indoor environments; thereby producing more complete maps that can be directly used for robotic navigation tasks. The performance of our framework is validated using real-world data.
\end{abstract}

\section{Introduction} \label{S:introduction}
In recent years, depth sensors have become a core component in a variety of robotic applications, including scene reconstruction, exploration, and inspection. However, commodity-grade RGB-D cameras, such as Microsoft Kinect and Intel RealSense, suffer from limited range and produce images with noise and missing data in view of surfaces that are too shiny, glossy, bright, or simply too far away. In robotic scenarios, this may lead to inefficient and inaccurate mapping performance when only the raw sensor data is used.

\edit{This paper studies the problem of \textit{depth completion} applied in the context of \textit{robotic mapping}.
Our goal is to create more complete spatial maps of cluttered 3D environments for robotic navigation purposes.
This is achieved by filling in holes found in raw depth images that are used for mapping.}

Recently, several deep learning-based approaches for depth completion using RGB-D images have been proposed \citep{Zhang2018,Huang2019} which effectively use colour information to enhance depth. However, propagating the completed depth into robotic frameworks remains an open challenge. A key issue is associating the completed areas with reliable measures of depth uncertainty, such that they can be used as an input for probabilistic mapping. Though several works have tackled uncertainty estimation for depth completion, they do not address using this information for 3D reconstruction \citep{Kendall2017} and largely focus on LiDaR-based sensors in outdoor environments \citep{Teixeira2020,Eldesokey2020a,Eldesokey2020b}. 

\begin{figure}[!t]
    \centering
    \includegraphics[width=0.45\textwidth]{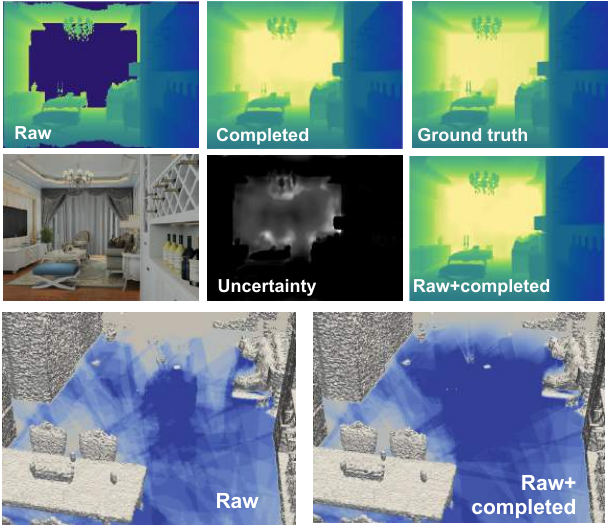}
    \caption{Overview of our approach for mapping with depth completion. \textit{Top}: Our depth completion network takes raw RGB-D images to predict the completed depth and depth uncertainty, which are used as additional inputs for probabilistic 3D mapping. \textit{Bottom}: By leveraging the network completions to complement the original raw depth data (right), we obtain more complete maps of free space when compared against using the raw depth alone (left). Darker shades of blue indicate areas of lower occupancy probability.
    } \label{F:teaser}
\end{figure}

To address this, we propose a new pipeline for mapping with \edit{probabilistic} depth completion; thus bridging the gap between computer vision algorithms and robotic applications. Inspired by the methods of \citet{Huang2019}, we introduce a network architecture that jointly predicts \edit{both} depth and depth uncertainty from RGB-D images by leveraging principles of Bayesian deep learning. Our approach exploits the processed images online as an additional input in the occupancy-based volumetric framework of \citet{Vespa2018} and \citet{Funk_arXiv2020}. This procedure enables us to produce maps with more discovered \textit{obstacle-free space} in the environment compared to using the raw images alone, as visualised in \Cref{F:teaser}, which is \edit{necessary} for robotic navigation tasks \edit{in initially unknown environments \citep{Funk_arXiv2020,Fehr2019}}.

The contributions of this work are:
\begin{enumerate}
    \item A new deep learning architecture providing uncertainty estimates for \edit{the probabilistic} depth completion of RGB-D images.
    \item The integration of our network in the volumetric mapping framework of \citet{Vespa2018} and \citet{Funk_arXiv2020}. We use the completed depth images with the predicted depth uncertainties in online probabilistic occupancy mapping to obtain more complete free space maps for robotic navigation tasks.
    \item The extensive evaluation of our framework using synthetic and real-world \edit{data showcasing} its performance.
\end{enumerate}
We plan to open-source our network implementation for usage and further development by the community. 

\section{Related Work} \label{S:related_work}
Algorithms for depth estimation and spatial mapping play a key role in many robotic applications and are the subject of a large and growing body of research. In this section, we review previous studies most related to our work.

Traditional methods for depth completion adopt hand-crafted kernels or features to compute the missing values \citep{Ferstl2013,Xue2017}. More recent algorithms \citep{Eldesokey2020a,Eldesokey2020b,Teixeira2020,Zhang2018,Huang2019,Uhrig2017,Ma2018} exploit deep learning for improved performance and generalisation capabilities. Our work focuses on the task of \textit{guided} depth completion, where the goal is to predict the dense depth values at every pixel based on the raw depth and a paired colour image. \citet{Uhrig2017} propose a sparse convolution layer which explicitly handles missing data to allow for inputs with varying degrees of sparsity. In a similar problem setup, \citet{Ma2018} use an encoder-decoder network to combine RGB and depth information within the underlying feature space. Recently, \citet{Eldesokey2020b} present a network based on normalised convolution layers which supports very sparse depth inputs and also provides confidence measures for the depth predictions. However, the aforementioned studies focus on completing sparse LiDaR-based data in outdoor scenarios and are thus not applicable to the types of degradation obtained with commodity-grade RGB-D cameras, as considered in our work.

For hole-filling with RGB-D cameras, \citet{Zhang2018} exploit the encoder-decoder architecture using dense occlusion boundaries and surface normals predicted from the colour image as secondary features to aid depth completion. Their approach involves an expensive loss optimisation step, making it unsuitable for real-time mapping. Building upon their ideas, \citet{Huang2019} introduce a network with a self-attention mechanism and boundary consistency to improve completion accuracy and speed. We propose an extension of their architecture which also predicts the uncertainties in the completed depth.

While significant work has been done on depth completion in the 2D image plane, applying these concepts to 3D mapping in robotics is a relatively unexplored \edit{research area}. Recently, \citet{Teixeira2020} introduced a depth completion algorithm for real-time aerial robotic applications. Similar to us, they obtain probabilistic depth predictions by estimating pixelwise uncertainties. However, they consider LiDaR-based sensing and do not use the completed images for 3D mapping. Most resembling our work is the approach of \citet{Fehr2019}, which uses an augmented depth sensor based on sparse inputs for robotic navigation. They show that their system uncovers more free space in \edit{unknown environments} when compared against using raw depth alone, \edit{thereby improving} planning performance. Although our work shares the same motivation, a key difference is that, instead of feeding the completed depth directly into a dense mapping framework, we adopt a fully probabilistic strategy based on the depth uncertainties provided by our new modified network.

Uncertainty in depth completion is crucial as it provides a reliability measure for fusing new predicted measurements into the map. One approach is to exploit confidence as a process internal to deep learning \citep{Qiu2019} to obtain more accurate dense depth outputs, i.e. by leveraging uncertainty as a weight map within the network architecture. An alternative is to treat uncertainty as an auxiliary \edit{network output} to obtain pixelwise uncertainties \citep{Kendall2017} or confidence maps \citep{Eldesokey2020a,Eldesokey2020b,Teixeira2020}. We follow the second class of approaches to extract explicit uncertainty values as inputs for mapping. Although, like us, \citet{Kendall2017} learn uncertainty in depth regression problems, to our knowledge, no prior work has applied these ideas in the context of probabilistic robotic mapping.

Another line of work focuses on volumetric scene completion directly in 3D space. For example, \citet{Song2017} predict volumetric occupancy and semantic labels from a single-view depth map. \citet{Dai2018} complete 3D geometry with per-voxel semantic labels from partial scans. However, as these methods require significant computational processing, they are not viable for real-time, online applications.

\section{Approach} \label{S:approach}
\begin{figure}[!h]
    \centering
    \includegraphics[width=0.48\textwidth]{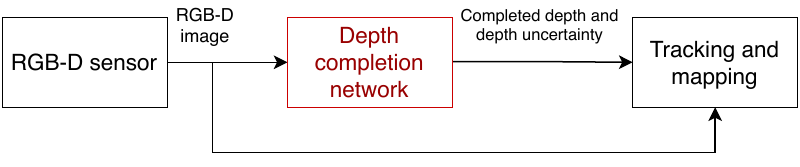}
    \caption{Overview of our proposed approach. We leverage a network for depth completion with uncertainty to improve the input for probabilistic tracking and mapping.}
    \label{F:overview}
\end{figure}

In this section, we propose a new approach for tracking and mapping using completed depth images with predicted depth uncertainty. A system overview is depicted in \Cref{F:overview}. As shown, we process the raw images from a RGB-D camera using a \edit{probabilistic} depth completion pipeline online to improve the input for \ac{SLAM}. Note that, while our approach is applicable for any \ac{SLAM} scenario, in this paper, we focus on \textit{mapping} only to show improvements for \edit{free space mapping} in unknown environments.
The following sub-sections describe our strategy for probabilistic depth completion before outlining the \ac{SLAM} framework.

\subsection{Network Architecture} \label{SS:network_architecture}

\begin{figure*}[!h]
    \centering
    \includegraphics[width=\textwidth]{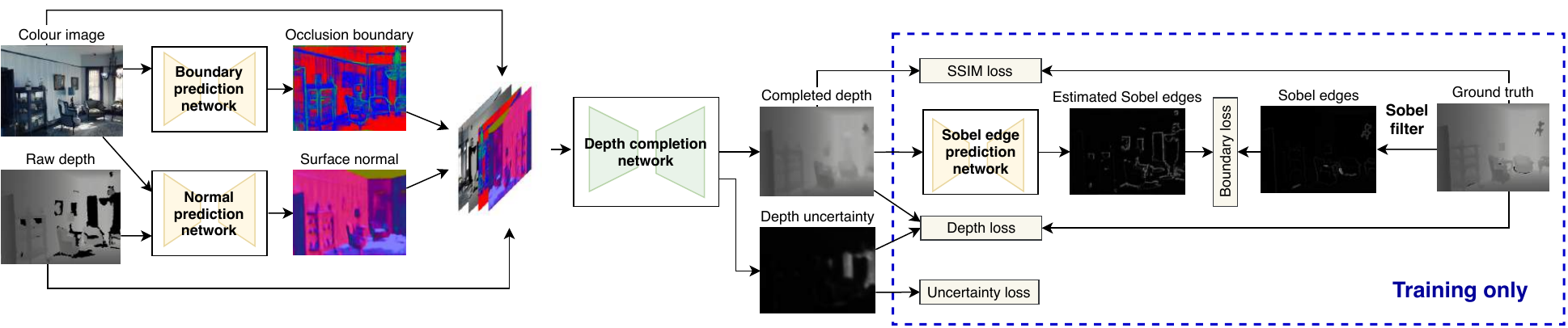} 
    \caption{Our \textit{probabilistic} depth completion system pipeline including the training framework with different training loss components (\Cref{SS:loss_function}). Given an input RGB-D image, we predict surface normals and boundaries and pass them to the probabilistic depth completion network (\Cref{F:architecture}) to predict depth and associated uncertainty. Black in the depth images indicates missing information.}
    \label{F:pipeline}
\end{figure*}

\begin{figure}[!h]
    \centering
    \includegraphics[width=0.48\textwidth]{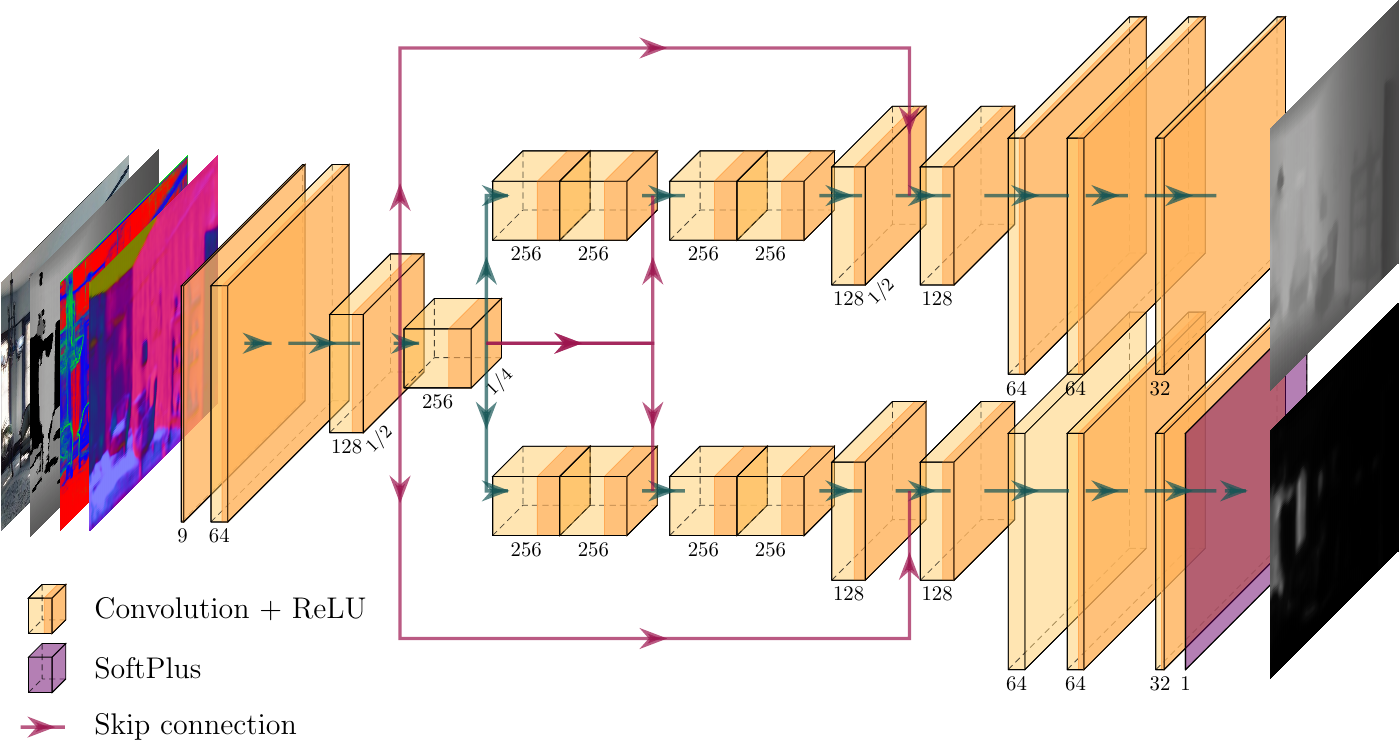}
    \caption{Our architecture for depth completion with uncertainty. We extend the network of \citet{Huang2019} with a second output decoder for uncertainty prediction. Our network takes as input raw RGB-D, surface normals and boundaries (left), and outputs completed depth (top-right) and pixelwise uncertainty (bottom-right).}
    \label{F:architecture}
\end{figure}

Our goal is to complete the depth channel of an RGB-D image and predict the associated pixelwise depth uncertainty, to be used as input for probabilistic tracking and mapping. To achieve this, we develop a pipeline based on the depth completion network proposed by \citet{Huang2019}. An overview of the depth completion sub-system is shown in \Cref{F:pipeline}. \Cref{F:architecture} details our new network architecture.

The main features of the depth completion network are the use of a self-attention mechanism and boundary consistency to produce depth maps with high quality and structure. Following \citet{Zhang2018}, we predict surface normals and occlusion boundaries from the raw RGB-D image and use them as additional input features to the network. To estimate surface normals, we employ the hierarchical RGB-D fusion network of \citet{Zeng2019}, which has state-of-the-art performance. Our boundary estimation network is based on the approach of \citet{Zhang2017}, using only RGB channels. The normals and boundaries are concatenated with the raw RGB-D image and used for the learning task.

To predict depth uncertainty, we leverage the Bayesian deep learning concepts of \citet{Kendall2017}. Our key idea is to introduce a second decoder on the \edit{network output} to learn the mapping to the input uncertainty in the completed depth. This is illustrated in the bottom branch of the architecture in \Cref{F:architecture}. We use a SoftPlus activation function (purple) to constrain the output uncertainty to be non-negative. By using a network with two different branches, the encoder of the original network captures the latent features common to the completed output depth and associated uncertainty, before they are processed separately to account for individual information. Moreover, we increase the number of channels per layer to 64 from 48 in the original network to provide a larger latent space for learning in the dual prediction task.

\subsection{Loss Function} \label{SS:loss_function}
We assume a Gaussian likelihood to model our aleatoric uncertainty \citep{Kendall2017}. Our loss function for depth completion with uncertainty is then the weighted sum of errors:
\begin{equation}
\begin{split}
    L ={}& \frac{1}{N}\sum_{p=1}^N \frac{1}{\sigma(\mathbf{x}_p)^2} |y_p-f(\mathbf{x}_p)|^2
    + \log (\sigma(\mathbf{x}_p)^2) \\
    &+ \lambda_{\mathrm{BC}} |y_{\mathrm{Sobel},p}-f_{\mathrm{Sobel}}(f(\mathbf{x}_p))| \\
    &+ \lambda_{\mathrm{SSIM}}SSIM(y_p,f(\mathbf{x}_p)) \, \mathrm{,} \label{E:loss}
\end{split}
\end{equation}
where $N$ is the number of pixels $p$ in an image, $\mathbf{x}_p$ is the input vector of features for the depth completion network (raw depth, RGB, estimated normals and boundaries), $y_p$ is the ground truth depth, $f(\cdot)$ and $\sigma(\cdot)$ are the completed depth and associated uncertainty output by the network, respectively, following directly from the negative log likelihood assuming Gaussian uncertainty, and $\lambda_{\mathrm{BC}}$ and $\lambda_{\mathrm{SSIM}}$ are tunable parameters. 
We only consider pixels in areas \textit{observed}, i.e. non-zero, in the ground truth depth image.

Following \citet{Huang2019}, we also include \edit{a boundary related loss (third term) to enforce boundary consistency and a structural related loss based on the \ac{SSIM} measure \citep{Wang2004} to reduce distortion and enhance structural quality (fourth term)}. The \edit{former} is computed by training a model to learn the Sobel edges associated with the completed depth supervised by those computed from the ground truth depth. The different components of the loss function are depicted in the dashed blue box in \Cref{F:pipeline}.

\subsection{Mapping} \label{SS:mapping}

We use an occupancy map to model the environment, as this representation is suitable for integrating noisy sensor measurements and explicitly captures free space for robotic planning applications.
Our approach leverages the multi-resolution occupancy mapping (`MultiresOFusion') and dense volumetric \ac{SLAM} framework from \citet{Funk_arXiv2020}.
This pipeline is an extension of \textit{supereight} \cite{Vespa2018} that enables integrating data at multiple octree levels and explicitly maps free space, while \edit{performing significantly better than other occupancy mapping frameworks, as shown in \citep{Funk_arXiv2020}.}

\begin{figure}[htb!]
    \centering
    \includegraphics[ width=0.97\columnwidth]{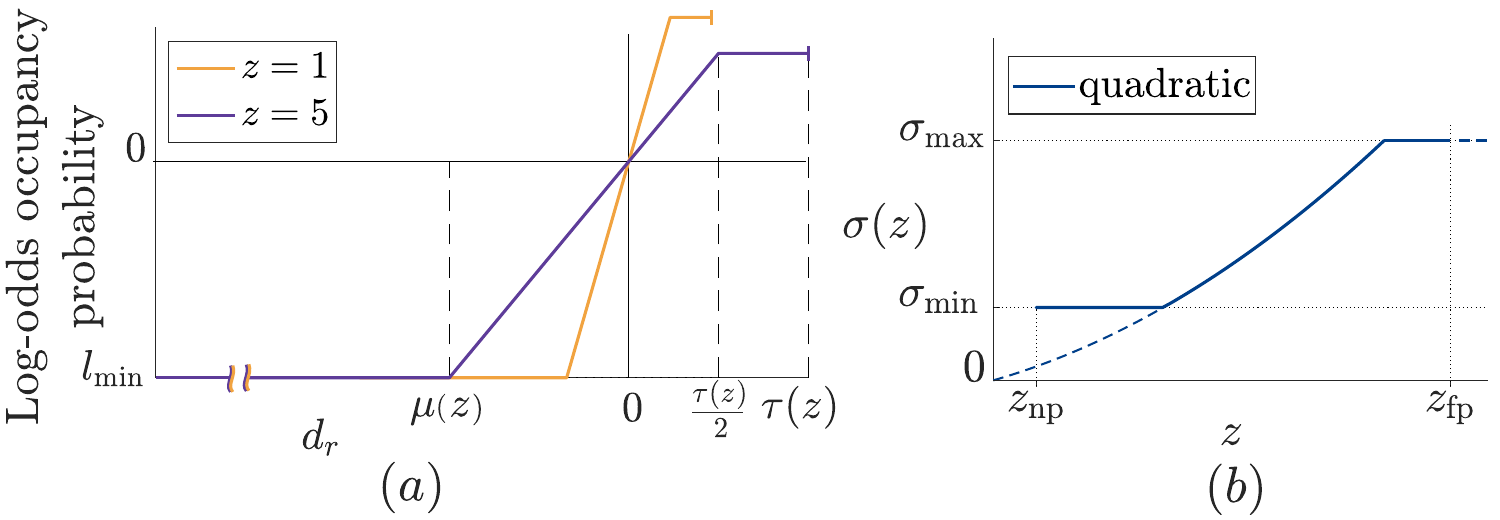}
    \caption{\edit{(a) Inverse sensor model for fusing new data into a map. Occupancy probability as a function of the difference $d_r$ from a query point to depth measured along a ray. 
    (b) Measurement uncertainty model for mapping with raw depth in \textit{supereight} `MultiresOFusion' \citep{Funk_arXiv2020}. Standard deviation $\sigma$ given depth measurement $z$.}}
    \label{F:supereight_sensor}
\end{figure}

To explain the role of depth uncertainty for mapping in our approach, we briefly overview the `MultiresOFusion' probabilistic inverse sensor model used to fuse new depth measurements into the map.
The inverse sensor model, inspired by \citet{Loop2016}, uses a piecewise linear function in log-odds space. The model produces probabilities expressed in log-odds directly to match the representation of occupancy probabilities in the map. Given a noisy depth measurement $z$, we assume its standard deviation is:
\begin{equation}
    \sigma(z) = \min\left( \max\left( k_{\sigma} {z}^2, \ \sigma_{\min} \right), \ \sigma_{\max} \right),
    \label{E:sigma}
\end{equation}
as shown in the right plot in \Cref{F:supereight_sensor}, where $k_{\sigma}$, $\sigma_{\min}$ and $\sigma_{\max}$ are constants.
This corresponds to a triangulation-based depth camera noise model.
\edit{The inverse sensor model is used to compute the log-odds occupancy probability given the distance $d_r$ from a query point to the measured depth $z_r$ along the ray as shown in the left plot. Log-odd values in front of the surface are clipped at $l_{\min}$ reached at $\mu = 3 \sigma$ and gradually increases with distance, peaking halfway through the surface thickness $\tau(z)$.
Surface thickness is computed as:}
\begin{equation}
    \tau(z) = \min\left( \max\left( k_{\tau} z, \ \tau_{\min} \right), \ \tau_{\max} \right)\,,   
    \label{E:tau}
\end{equation}
where $k_{\tau}$, $\tau_{\min}$ and $\tau_{\max}$ are constants.
No voxels beyond $z_r + \tau$ from the camera are updated.
Larger values of $\sigma$ result in a more gradual increase of occupancy probability.
\edit{For fusing multiple measurements, we use a clamped accumulation log-odd occupancy as described by \citet{Vespa2018}}.

Our aim is to exploit the completed depth and depth uncertainty provided by our network to complement the raw sensor data captured by this model and improve mapping performance.
Specifically, we propose using the network-generated depth uncertainty instead of the measurement uncertainty above for completed depth areas.

\section{Experimental Results} \label{S:results}
This section presents our experimental results. First, we validate our probabilistic depth completion network via an ablation study. We then evaluate our 3D mapping framework in indoor environments using synthetic and real-world data.

\begin{table*}[h]
\centering
\setlength{\tabcolsep}{2.2pt}
\begin{tabular}{lccccccccc}
    Model           & {RMSE (m) $\downarrow$}    & {AUSE $\downarrow$}  & {\edit{SSIM $\uparrow$}}  & {MAE (m) $\downarrow$} & {$1.05 \edit{(\%)} \uparrow$} & {$1.10 \edit{(\%)} \uparrow$} & {$1.25 \edit{(\%)} \uparrow$} & {$1.25^2 \edit{(\%)} \uparrow$} & {$1.25^3 \edit{(\%)} \uparrow$} \\ \midrule[1.2pt]
    (i) 2 decoders (64) (ours)   & $\mathbf{0.3154}$        & 0.1849  & $\mathbf{\edit{0.9054}}$ & 0.1282 & $\mathbf{85.34}$ & $\mathbf{90.01}$ & 93.92 & 96.55 & 97.85 \\ \midrule
    (ii) 1 decoder (48)  & 0.3484        & $\mathbf{0.1771}$    & \edit{0.8820} & 0.1538 & 80.19 & 85.08 & 90.55 & 95.16 & 97.22 \\ \midrule
    (iii) 2 decoders (48)    & 0.3187        & 0.2115 & \edit{0.8994} & 0.1335   & 85.22 & 89.61 & 93.52 & 96.29 & 97.68 \\ \midrule
    (iv) 2 decoders (48), depth weights from \citep{Huang2019}   & 0.3166 & 0.1996  & \edit{0.9034}   & $\mathbf{0.1268}$ & 80.30 & 88.69 & $\mathbf{94.20}$ & $\mathbf{96.89}$ & $\mathbf{98.02}$
\end{tabular}
\caption{Comparison of our proposed 2-decoder, 64-channel network for depth completion with uncertainty (top) against benchmarks derived from \citet{Zhang2018} on the Matterport3D test set. Our architecture achieves good depth uncertainty measures without compromising depth prediction accuracy. The number of channels per layer in the network is in parentheses.} \label{T:results_ablation}
\end{table*}

\begin{figure*}[!h]
  \centering
  \includegraphics[width=\textwidth]{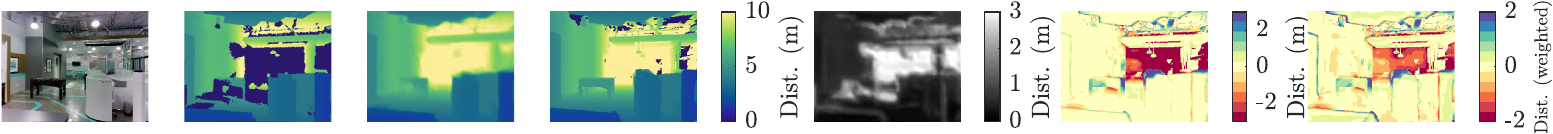}
  
  \vspace{0.4mm}
  \includegraphics[width=\textwidth]{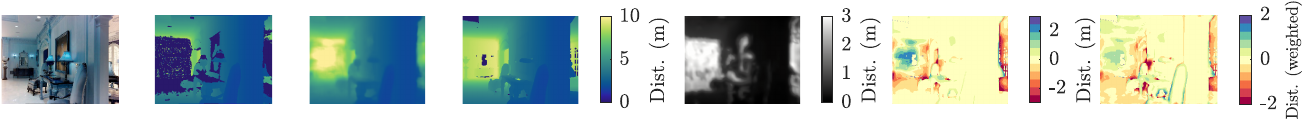}

  \vspace{0.4mm}
  \includegraphics[width=\textwidth]{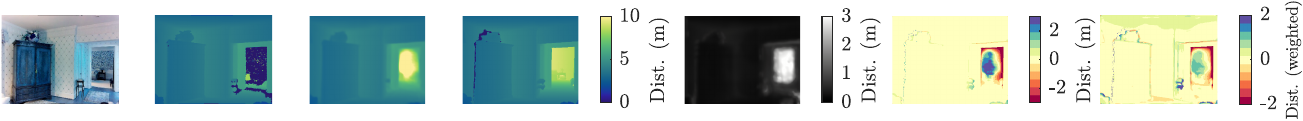}

  \vspace{0.4mm}
  \includegraphics[width=\textwidth]{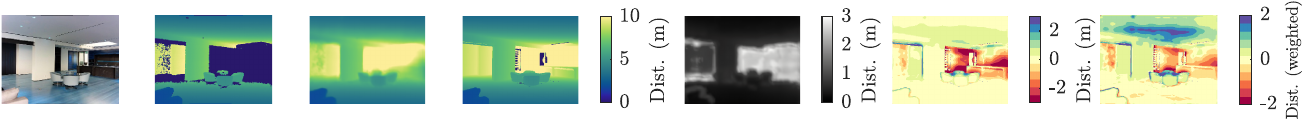}
\caption{Examples of our proposed 2-decoder, 64 channel per layer \edit{probabilistic} depth completion network outputs on images from different sequences from the Matterport3D test set. \textit{Left to right}: RGB, raw depth, completed depth, ground truth depth (obtained from \citep{Zhang2018}), depth uncertainty (standard deviation), depth error, depth error weighted by the standard deviation. We validate that our network completes the holes in the raw depth images and provides higher uncertainty estimates in areas where data is missing.
} \label{F:matterport_results_completion}
\end{figure*}

\subsection{Training Procedure} \label{SS:training_procedure}
We trained our depth completion system end-to-end on Matterport3D, a large-scale RGB-D dataset \citep{Chang2017} representative of an indoor exploration task. For training, (complete) ground truth depth was obtained from \citet{Zhang2018} based on multi-view reconstruction. Unless otherwise specified, 129816 and 36252 images from the dataset were used for training and testing, respectively, with a resolution of $320$\,px\,$\times$\,$240$\,px. We used the Adam optimiser with a weight decay of $10^{-3}$, a learning rate of $10^{-5}$, and set $\lambda_{SSIM} = \lambda_{BC} = 1$ in \Cref{E:loss}. The models were implemented in PyTorch and training was done on a NVIDIA GeForce RTX 2080 Ti GPU with a 3.9GHz AMD Ryzen 9 3900X CPU. On this machine, one forward pass through the pipeline takes $\sim0.2$\,s.

\subsection{Ablation Study} \label{SS:ablation_study}

Our first aim is to evaluate our new two-decoder network for depth completion with uncertainty. To this end, an ablation study is conducted to investigate the benefits of separating the two outputs in the proposed architecture and training the model end-to-end for both depth completion and uncertainty. We compare: (i) our proposed architecture with two decoders (\Cref{F:architecture}); (ii) the original architecture of \citet{Huang2019} simply extended with a single shared output decoder for depth and uncertainty; (iii) a smaller variant of our architecture, using 48 channels per layer as in \citep{Huang2019} instead of 64; and (iv) the same architecture as in (iii), but freezing the weights of the encoder and depth completion decoder parts using the trained model in \citep{Huang2019}, such that only the uncertainty output features are learned. Apart from (iv), we initialised each model with random weights, then let it train 
and report the best epoch. We also experimented with initialising (iii) using the pretrained weights from (iv) for training and confirmed that this has no significant impact on the final results, 
but does speed up convergence.

To evaluate prediction accuracy, we consider the \ac{RMSE}, the \ac{MAE}, the percentages of predicted pixels $\mathbf{p}_\mathrm{pred}$ within an interval $\delta = \frac{|\mathbf{p}_\mathrm{pred}-\mathbf{p}_\mathrm{true}|}{\mathbf{p}_\mathrm{true}}$, where $\mathbf{p}_\mathrm{true}$ are the corresponding pixel values obtained from the ground truth image and $\delta \in \{1.05,\,1.10,\,1.25,\,1.25^2,\,1.25^3\}$, \edit{and the \ac{SSIM}}. To evaluate the quality of the uncertainty, we measure the \ac{AUSE}. As explained by \citet{Ilg2018}, this metric captures the correlation between the estimated uncertainty and prediction error. Our \ac{AUSE} values are computed based on all pixels in the test set.

\Cref{T:results_ablation} summarises our results. With the lowest \ac{AUSE}, the single-decoder network produces the best uncertainty measure at the cost of lowest prediction accuracy. In contrast, training \textit{only} a second decoder yields low error, but poor uncertainty, as the shared encoder weights are fixed and optimised for the depth completion problem. By increasing the model latent space and training the network end-to-end, our proposed model obtains relatively low \ac{AUSE} without compromising prediction accuracy with respect to the original implementation. For visual validation, \Cref{F:matterport_results_completion} presents example results of our proposed larger 2-decoder model on the Matterport3D test set. This way, we achieve high-quality completed depth with reliable, and, importantly, consistent uncertainty estimates, which we exploit in the next sub-sections to improve probabilistic mapping performance.

\subsection{Evaluation on Synthetic Data} \label{SS:interiornet}
We perform a quantitative evaluation of our approach for occupancy mapping using trajectory sequences from the synthetic RGB-D dataset InteriorNet \citep{Li2018}, in which ground truth depth images and pose data are available. Our aim is to show that mapping using the network predicted depth and uncertainty leads to more complete final maps and a greater volume of free space discovered in the environment, which is a key requirement for safe robotic planning and navigation.

The ground truth depth from InteriorNet contains no measurement noise.
To simulate realistic noisy depth images, we degrade the ground truth following the quadratic noise model for the Kinect sensor developed by \citet{Nguyen2012}:
\begin{equation}
\sigma_z(z) = 0.0012 + 0.0019 \, (z-0.4)^2 \, \mathrm{,}
\end{equation} \label{E:kinect_noise}
where $\sigma_z$ is the standard deviation of lateral noise in metres at a pixel and $z$ is the corresponding depth measurement in metres. Additionally, a Gaussian filter with a standard deviation of $0.5$\,px is applied on the depth image to blur the noise between adjacent pixels.

The ground truth depth does not contain occlusion holes or missing depth measurements.
To create missing data for depth completion, we set practical sensing limits of $0.8$\,m$-6$\,m beyond which depth readings are zero. To generate occlusion holes and remove measurements based on the context and structure of the scene within this range, e.g. on textureless/reflective surfaces, we use the generative adversarial framework of \citet{AtapourAbarghouei2019}, which learns to predict depth holes from RGB. This network is trained on \edit{our Matterport3D training set}, since it contains real RGB/raw depth image pairs. We then train CycleGAN, an unpaired image-to-image translation network \citep{Zhu2017} to learn the visual domain shift between Matterport3D and InteriorNet using 10000 random images from each dataset, and apply the hole predictor on InteriorNet RGB images translated to the Matterport3D style. This procedure allows us to transfer the learnt structures of real holes to the synthetic dataset and thus generate realistically corrupted depth images.

\begin{figure*}[!h]
    \centering
    \includegraphics[height=0.012\textwidth]{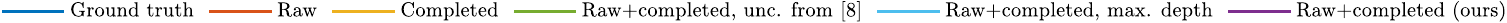}
    \vspace{1mm}
    
    \begin{subfigure}[h]{0.323\textwidth}
    \includegraphics[height=0.11\textheight]{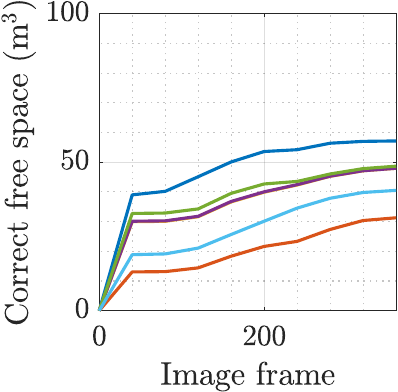}
    \includegraphics[height=0.11\textheight]{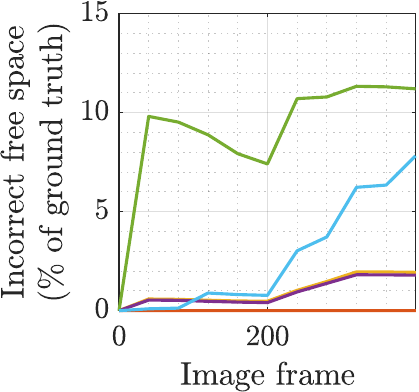}
    \caption{Seq. 1} \label{SF:results_in_space1}
    \end{subfigure}
    \begin{subfigure}[h]{0.323\textwidth}
    \includegraphics[height=0.11\textheight]{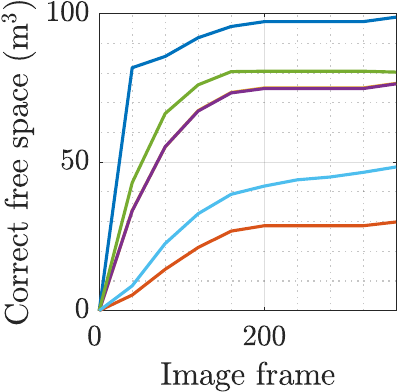}
    \includegraphics[height=0.11\textheight]{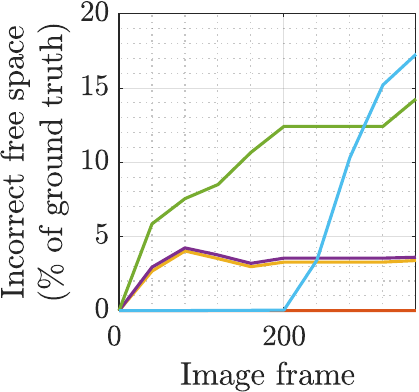}
    \caption{Seq. 2} \label{SF:results_in_space2}
    \end{subfigure}
    \begin{subfigure}[h]{0.323\textwidth}
    \includegraphics[height=0.11\textheight]{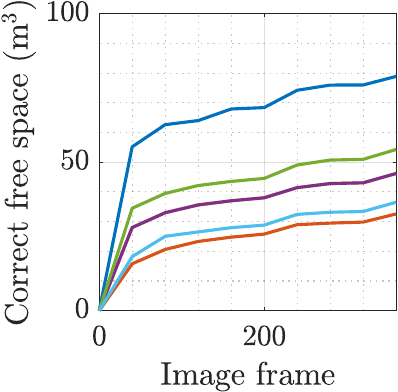}
    \includegraphics[height=0.11\textheight]{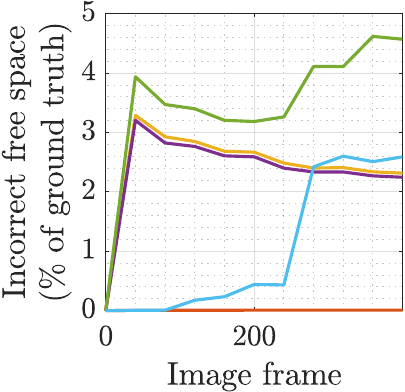}
    \caption{Seq. 3} \label{SF:results_in_space3}
    \end{subfigure}
    \caption{Comparison of correct and incorrect free space volume discovered during three trajectory sequences from InteriorNet using different depth and uncertainty inputs for mapping. Mapping with \edit{probabilistic} depth completion leads to more correct free space mapped throughout the image sequence with relatively small errors. Note the different scales on the $y$ axes for correct and incorrect free space.} \label{F:results_interiornet_space}
\end{figure*}

For depth completion, we use our trained 2-decoder, 64-channel network from \Cref{SS:ablation_study}, further fine-tuned on the corrupted InteriorNet depth data with 110000 and 28000 images for training and testing, respectively. The images are downsampled to $320\,$px$\times\,240$\,px and we apply the same optimisation algorithm as detailed in \Cref{SS:training_procedure}.

For spatial mapping, we use \textit{supereight} `MultiresOFusion' \citep{Funk_arXiv2020} with a voxel resolution of $0.0146$\,m in a $15$\,m\,$\times\,15$\,m\,$\times\,15$\,m volume. We use the inverse sensor model in \Cref{F:supereight_sensor} to integrate raw depth data into the map, setting the constants $k_{\tau}$, $\tau_{\min}$, and $\tau_{\max}$ as $0.026$, $0.06$\,m, and $0.16$\,m, respectively. To capture measurement uncertainty, we consider the quadratic uncertainty model in \Cref{E:sigma} with $k_{\sigma} = 0.0016$\,m, $\sigma_{\min} = 0.005$\,m, and $\sigma_{\max} = 0.02$\,m resulting in $\sigma_r = 0.005$\,m at $z_r = 1$\,m for the raw depth.

We map three trajectories from different InteriorNet scenes not present in the training set\footnote{Seq. 1: `3FO4K9H4NDAO (7)'; Seq. 2: `3FO4JVHRJC4T (7)'; Seq. 3: `3FO4JXILITSO (7)'. Trajectory numbers are given in the parentheses.}. We use $400$ images for mapping per sequence, picking large rooms with wide ranges of motion to highlight the advantages of applying depth completion when data is missing. Our experiments compare: (i) the raw depth images with a quadratic depth uncertainty as given in \Cref{E:sigma} (R); (ii) the completed images with the predicted depth uncertainty from our network (C); and a combination of the two (R+C), where the raw depth is used in known areas, and the \edit{invalid depth pixels are completed}. \edit{As baselines with completion, we study (iii) using the quadratic sensor model in \textit{both} the raw and network completed areas (R+C, unc. from \citep{Funk_arXiv2020}) and (iv) simply filling in the invalid areas with the maximum camera range ($8$\,m) and depth uncertainty $3\sigma_r$ conservatively set to this value (R+C, max. depth).
Our proposed approach (v) is using the detailed raw depth with the quadratic sensor model in valid areas and depth completion with the network predicted uncertainty to fill in the rest (R+C (ours)).
This method is depicted in the top images in \Cref{F:teaser} and the system diagram in \Cref{F:overview}.
In (ii) and (v),} for pixels where the network-generated depth is used, we compare the network depth uncertainty with the quadratic uncertainty model (\Cref{E:sigma}) when using the network completed depth. If the network depth uncertainty is more than 2 times greater than the quadratic uncertainty \textit{only} free space is integrated for this pixel, otherwise normal integration is performed. This prevents us from creating incorrect surfaces for very uncertain completions while still obtaining usable probabilistic free space estimates.

\begin{table}[h]
\setlength{\tabcolsep}{2.6pt}
\centering
\begin{tabular}{llccc}
\makecell{Seq.} & \makecell{Method} & \makecell{Correct free \\ space (m$^3$) $\uparrow$} & \makecell{Incorrect free \\ space (m$^3$) $\downarrow$} & \makecell{Mesh \\ accuracy (m) $\downarrow$} \\ \midrule[1.2pt]
 \multirow{5}{*}{1}  & R                                                  & \edit{31.6748} & \edit{0.0092} & \edit{0.0891}    \\
                     & C                                                  & \edit{48.2372} & \edit{1.1175} & \edit{0.1652}    \\
                     & \edit{R+C, unc. from \citep{Funk_arXiv2020}}       & \edit{48.9464} & \edit{6.4669} & \edit{1.0337}    \\
                     & \edit{R+C, max. depth}                             & \edit{40.8428} & \edit{4.5132} & \edit{0.3379}    \\
                     & R+C (ours)                                         & \edit{48.2307} & \edit{1.0351} & \edit{0.2008}    \\ \midrule
 \multirow{5}{*}{2}  		     & R                                                  & \edit{32.1136} & \edit{0.0066} & \edit{0.0884}    \\
                     & C                                                  & \edit{77.9038} & \edit{3.7560} & \edit{0.2073}    \\
                     & \edit{R+C, unc. from \citep{Funk_arXiv2020}}       & \edit{81.5002} & \edit{15.6306} & \edit{1.1153}\\
                     & \edit{R+C, max. depth}                             & \edit{47.8514} & \edit{17.4437} & \edit{0.2942}\\
                     & R+C (ours)                                         & \edit{77.7847} & \edit{3.9155} & \edit{0.2939}    \\ \midrule
 \multirow{5}{*}{3}   & R                                                  & \edit{34.0504} & \edit{0.0070} & \edit{0.0704}    \\
                     & C                                                  & \edit{47.6763} & \edit{1.8470} & \edit{0.0885}    \\
                     & \edit{R+C, unc. from \citep{Funk_arXiv2020}}       & \edit{55.6404} & \edit{3.6456} & \edit{0.6194}\\
                     & \edit{R+C, max. depth}                             & \edit{37.8850} & \edit{2.0651} & \edit{0.3551}\\
                     & R+C (ours)                                         & \edit{47.5621} & \edit{1.7918} & \edit{0.1725}    \\ \midrule
\end{tabular}
\caption{Evaluation of different depth and uncertainty inputs for mapping using three sequences from InteriorNet. `R', `C' and `R+C' represent raw depth, completed depth, and a combination of the two. Depth completion enables mapping more free space without significantly compromising reconstruction accuracy.} \label{T:results_interiornet}
\end{table}
Our evaluation metrics are the volumes of correct and incorrect mapped free space in the environment with respect to the map generated with ground truth depth at a given image frame. Voxels with occupancy probability $<0.04\%$ are considered to be free in the reconstructions; for ground truth, we use a less conservative threshold of $<3\%$. 
\edit{These thresholds can be tuned to balance between extra free space exploration and false-positive free space in a given scenario.}
To measure accuracy in the final reconstructions, we create meshes using marching cubes, and compute the average distance from the ground truth mesh to an output mesh.

The evaluation results are summarised in \Cref{T:results_interiornet}. Using \edit{depth completion} in our mapping pipeline leads to remarkably more discovered free space volume since, thanks to the predictions, we can capture the free space associated with \textit{all} pixels, instead of only those with valid raw depth measurements. \edit{The two `R+C' baselines produce high volumes of incorrect free space with their simple heuristics. In contrast, using our probabilistic network output yields much smaller inaccuracies relative} to the gain in correct free space.
Our proposed combined approach (`R+C') \edit{has the benefit of preserving} real, \edit{detailed} raw depth where it is available. Finally, though the reconstruction accuracy with completion is slightly worse when compared to using the raw depth alone (`R'), it is not drastically degraded with respect to the size of the rooms. We emphasise here that our aim is \textit{not} to achieve higher-quality fine-scale \edit{surface} reconstructions, but rather \edit{create a free space map of} the environment suitable for motion planning while preserving its \edit{global} structure.

\begin{figure}[!h]
    \centering
    \includegraphics[width=0.99\columnwidth]{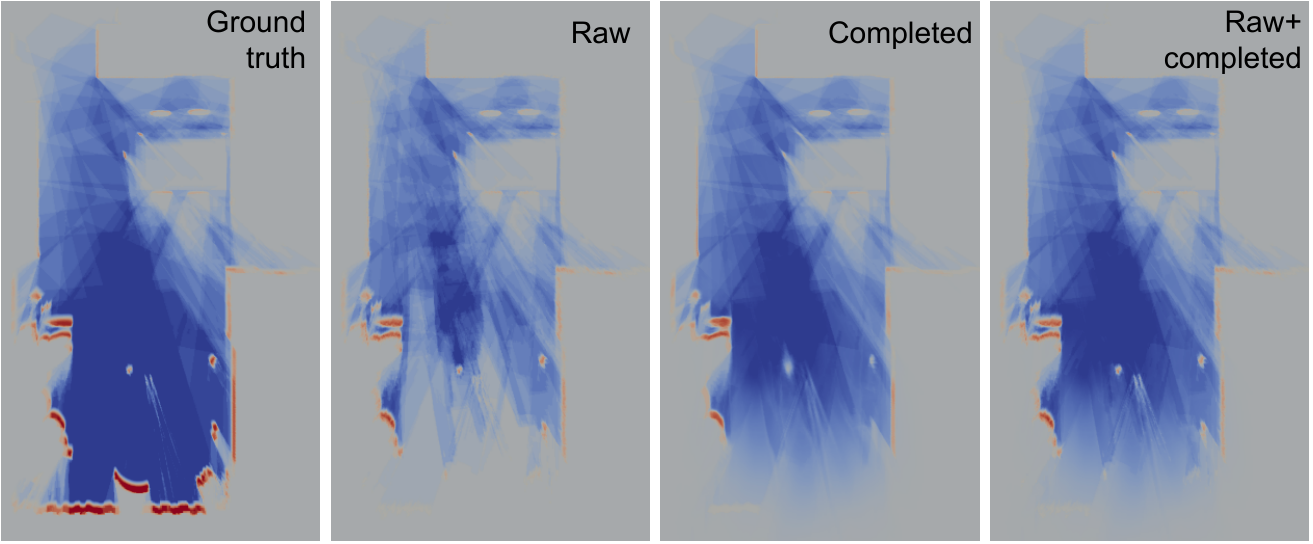}
    \caption{Comparison of occupancy map cross-sections for Seq. 1 of InteriorNet at image frame $160$. Blue to red colours encode occupancy probability (cream is unknown). Depth completion enables mapping more free space (blue) on the side of the room opposite the depth camera (bottom), which is outside the measurement range.
    } \label{F:results_interiornet_slices}
\end{figure}

\Cref{F:results_interiornet_space} depicts the evolution of the mapped free space during the three sequences using \edit{different mapping strategies}. As a qualitative result, \Cref{F:results_interiornet_slices} illustrates occupancy map cross-sections at image frame $160$ of Seq. 1. The plots in \Cref{F:results_interiornet_space} verify that the completion methods consistently map significantly more free space compared to using only the raw depth (orange), even with a small number of images, while free space error is relatively low and grows slowly. \Cref{F:results_interiornet_slices} depicts visually the greater proportion of free space (blue areas) achieved using depth completion, especially on the side of the room away from the depth camera (bottom). This portrays the benefit of using our framework in large environments where raw depth coverage is limited.

The bottom images of \Cref{F:teaser} show the occupancy map cross-sections from \Cref{F:results_interiornet_slices} overlaid on the output meshes obtained using the `R' and `R+C' approaches. We confirm that the free space in the room is much more complete using our depth completion pipeline. \edit{As expected, reconstruction quality remains visually similar; our strategy for integrating free space with highly uncertain completions prevents creating artefacts which may compromise navigation safety}.

\subsection{Evaluation on Real-World Data} \label{SS:tum}
We demonstrate our pipeline for occupancy mapping with \edit{probabilistic} depth completion using \edit{five sequences from} the TUM RGB-D dataset \citep{Sturm2012}, which \edit{contain} trajectory ground truth and RGB-D images captured with a Microsoft Kinect.
Note that TUM RGB-D does not include ground truth depth for evaluating accuracy as in \Cref{SS:interiornet}. Instead, the aim is to validate qualitatively the benefit of using our trained depth completion system to map free space using real images.

For mapping, we use \textit{supereight} `MultiresOFusion' with a $0.0146$\,m voxel resolution in a $15$\,m\,$\times\,15$\,m\,$\times\,15$\,m volume. The constants $k_{\tau}$, $\tau_{\min}$, and $\tau_{\max}$ are $0.05$, $0.06$\,m, and $0.16$\,m, respectively, and the quadratic uncertainty model in \Cref{E:sigma} uses $k_{\sigma} = 0.0025$\,m, $\sigma_{\min} = 0.0098$\,m, and $\sigma_{\max} = 0.0294$\,m. These parameters correspond to those used by \citet{Funk_arXiv2020} to evaluate `MultiresOFusion' on real-world data.
For depth completion, we use our 2-decoder, 64-channel network from \Cref{SS:ablation_study} trained on Matterport3D with $320\,$px$\times\,240$\,px images. Note that the weights finetuned on InteriorNet in \Cref{SS:interiornet} produced similar results.

\begin{figure}[!h]
  \centering
  \includegraphics[width=\columnwidth]{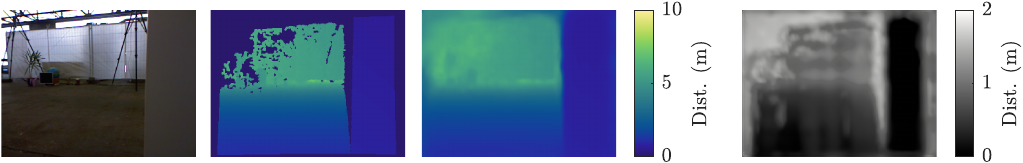}

  \includegraphics[width=\columnwidth]{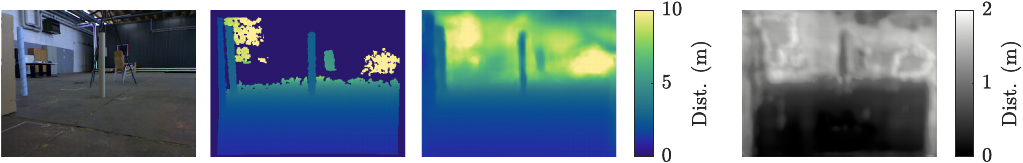}
  
  \includegraphics[width=\columnwidth]{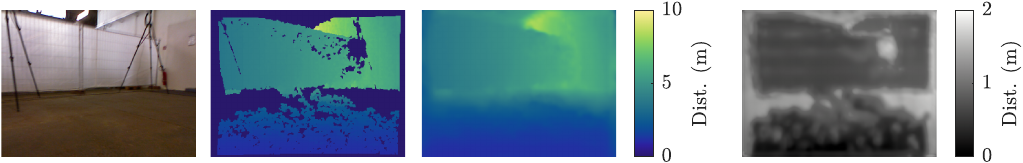}

  \caption{Examples of our \edit{probabilistic} depth completion network outputs on various images from TUM RGB-D \texttt{fr2/pioneer\_slam2}. \textit{Left to right}: RGB, raw depth, completed depth, depth uncertainty (standard deviation). Our network completes holes in the raw depth and provides valid uncertainty estimates for free space mapping.} \label{F:results_tum_completion}
\end{figure}
\begin{table}[!h]
\footnotesize
\centering
\begin{tabular}{ll|cccc}
     & & \multicolumn{4}{c}{\makecell{Discovered free space volume (m$^3$) \\ at sequence completion (\%)}} \\ \midrule
    \makecell[l]{Sequence} & \makecell{Method} & 25 (\%) & 50 (\%) & 75 (\%) & 100 (\%) \\ \midrule[1.2pt]
    \multirow{2}{*}{\makecell[l]{\texttt{fr1/} \\ \texttt{360}}}
                         & R      & 21.78 & 60.21 & 78.41 & 80.48    \\
                         & R+C    & 22.41 & 62.10 & 82.68 & 84.63    \\ \midrule
    \multirow{2}{*}{\makecell[l]{\texttt{fr2/} \\ \texttt{pioneer\_slam}}}
                         & R      & 108.25 & 239.18 & 289.55 & 306.33    \\
                         & R+C    & 170.08 & 313.39 & 368.51 & 382.46    \\ \midrule
     \multirow{2}{*}{\makecell[l]{\texttt{fr2/} \\ \texttt{pioneer\_slam2}}}
                         & R      & 93.28 & 159.02 & 245.67 & 273.20    \\
                         & R+C    & 103.50 & 202.45 & 326.58 & 354.35    \\ \midrule
    \multirow{2}{*}{\makecell[l]{\texttt{fr2/} \\ \texttt{pioneer\_slam3}}}
                         & R      & 87.40 & 116.11 & 212.29 & 268.37    \\
                         & R+C    & 142.21 & 185.44 & 291.43 & 348.24    \\ \midrule
     \multirow{2}{*}{\makecell[l]{\texttt{fr2/} \\ \texttt{pioneer\_360}}} 
                         & R      & 81.21 & 203.56 & 264.14 & 268.29    \\
                         & R+C    & 149.52 & 272.98 & 339.95 & 350.31
\end{tabular} \caption{\edit{Comparison of free space volumes (m$^3$) discovered during five sequences from TUM RGB-D using different depth and uncertainty inputs for mapping. Using the probabilistic network completed depth to complement raw depth (`R+C') yields faster free space mapping compared to using the raw depth alone (`R').}} \label{T:results_tum}
\end{table}

\begin{figure}[!h]    
   \vspace{1.2mm}
  
  \begin{subfigure}[h]{0.49\textwidth}
  \centering
  \includegraphics[height=0.345\columnwidth]{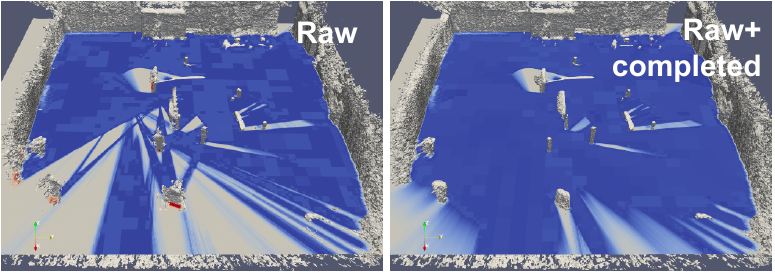}
  \end{subfigure}
\caption{\edit{Comparison of final occupancy map cross-sections and 3D meshes for TUM RGB-D \texttt{fr2/pioneer\_360}. Blue to red colours encode occupancy probability. Depth completion uncovers more free space while the surface reconstruction remains similar.}} \label{F:results_tum_space}
\end{figure}
We compare mapping using: raw depth with the quadratic sensor model in \Cref{E:sigma} (denoted by `R' in \Cref{SS:interiornet}) and our proposed combined approach (`R+C'), using the network-generated completed depth and depth uncertainty in \edit{invalid raw depth areas}. Examples of \edit{probabilistic} depth completion can be seen in \Cref{F:results_tum_completion}. As described in \Cref{SS:interiornet}, for our proposed method, we integrate only the free space for completed pixels where the network depth uncertainty is more than $2$ times greater than the quadratic uncertainty of the raw sensor model. As before, we measure free voxels based on an occupancy probability of $<0.04\%$.

\edit{\Cref{T:results_tum} shows the free space mapped during the \edit{sequences}. Similar to our InteriorNet experiments, using \edit{probabilistic} depth completion for mapping consistently yields faster free space discovery when compared against the raw data alone. The occupancy cross-sections in \Cref{F:results_tum_space} depict visually more free space (blue) at the end of the sequence using completion while the final 3D reconstruction remain similar and do not compromise global navigation safety.} These results validate our pipeline in real-world settings.

\section{Conclusions and Future Work} \label{S:conclusions}
This paper introduced a framework for volumetric mapping using depth completion with uncertainty. A core component of our pipeline is a new network architecture for jointly predicting missing depth and depth uncertainty based on images from commodity-grade RGB-D cameras in cluttered indoor environments. The probabilistic depth is used as an input for mapping to complement the raw depth images,  
allowing us to obtain more complete free space maps.

We performed an ablation study to validate our network for depth completion with uncertainty. The \edit{integrated} system for mapping with probabilistic depth was evaluated using synthetic RGB-D data. Our proposed approach using both raw and completed depth was shown to \edit{discover} correct free space \edit{most rapidly} without compromising map accuracy \edit{and safety in terms of false positive free space}.
This property is crucial for robotic planning \edit{in unknown environments}. Further tests validate our approach using real-world images.

One limitation is that our network completions are oversmoothed on depth discontinuities due to the high uncertainties present in these regions. Though our combined approach mitigates this issue by using raw depth data, in more complex environments, one could exploit the edge predictions available from training to preserve sharp boundaries. Another idea is to use recurrent networks to ensure consistency in depth prediction between consecutive images. Finally, we will extend our framework to active mapping problems.

\section*{Acknowledgement}
We would like to thank Xiao Gu for his advice on simulating realistic depth cameras and Dr. \v{Z}eljko Popovi\'{c} and Alexis Laignelet for helping with our experimental setup.

\bibliographystyle{IEEEtranN}
\footnotesize
\bibliography{2021-icra-popovic}

\end{document}